\documentclass{bmvc2k}

\title{Learning to Generate Novel Classes for Deep Metric Learning}

\usepackage[table]{xcolor}
\usepackage{xcolor}
\usepackage{enumitem}
\usepackage{soul}
\usepackage{tabularx}
\usepackage{xcolor}

\usepackage{amsfonts}
\usepackage{amsmath}
\usepackage{amssymb}

\usepackage{makecell}

\definecolor{green}{rgb}{0, 0.6, 0}
\definecolor{brown}{rgb}{0.65, 0.16, 0.16}

\addauthor{Kyungmoon Lee}{kyungmoon@postech.ac.kr}{1}
\addauthor{Sungyeon Kim}{sungyeon.kim@postech.ac.kr}{1}
\addauthor{Seunghoon Hong}{seunghoon.hong@kaist.ac.kr}{2}
\addauthor{Suha Kwak}{suha.kwak@postech.ac.kr}{1}

\addinstitution{
 POSTECH \\
 Pohang, South Korea
}
\addinstitution{
 KAIST \\
 Daejeon, South Korea
}

\runninghead{Lee et al.}{Learning to Generate Novel Classes for Deep Metric Learning}

\def\eg{\emph{e.g}\bmvaOneDot}

\def\ie{\emph{i.e}\bmvaOneDot}

\definecolor{forestgreen}{rgb}{0.13, 0.55, 0.13}

\begin{document}

\maketitle

\begin{abstract}
Deep metric learning aims to learn an embedding space where the distance between data reflects their class equivalence, even when their classes are unseen during training. However, the limited number of classes available in training precludes generalization of the learned embedding space. Motivated by this, we introduce a new data augmentation approach that synthesizes novel classes and their embedding vectors.
Our approach can provide rich semantic information to an embedding model and improve its generalization by augmenting training data with novel classes unavailable in the original data. We implement this idea by learning and exploiting a conditional generative model, which, given a class label and a noise, produces a random embedding vector of the class. Our proposed generator allows the loss to use richer class relations by augmenting realistic and diverse classes, resulting in better generalization to unseen samples. Experimental results on public benchmark datasets demonstrate that our method clearly enhances the
performance of proxy-based losses.
\end{abstract}

\section{Introduction}
\label{sec:introduction}
Deep metric learning is the task of learning an embedding space where data of the same class are placed closely so that the distance between data reflects their class equivalence. 
It has been a driving force behind recent advances in numerous computer vision and machine learning tasks including 
image retrieval~\cite{songCVPR16,kim2019deep,Sohn_nips2016}, face identification~\cite{Chopra2005,Schroff2015}, person re-identification~\cite{Chen_2017_CVPR}, and representation learning~\cite{kim2019deep,chen2020simple}. 
The main reason for adopting deep metric learning in these tasks is its generalization capability; the learned embedding space is expected to be well generalized to unseen classes so that it can be used to predict the class equivalence of a pair of data even when their classes are unavailable during training.

Various ways to improve the performance of deep metric learning have been studied so far, such as advanced loss functions~\cite{songCVPR16,Sohn_nips2016,wang2019multi,movshovitz2017no,kim2020proxy}, ensemble methods~\cite{opitz2018deep,ensemble_embedding}, regularization techniques~\cite{JACOB_2019_ICCV,mohan2020moving}, sample mining~\cite{sampling_matters,Harwood_2017_ICCV}, and sample generation~\cite{ko2020embedding,lin2018deep}.
Although the effectiveness of these methods has been demonstrated, we believe their generalization capability could be further improved in the sense that they are inherently limited only to classes available in a training set.

\begin{figure}
\begin{tabular}{cc}
\bmvaHangBox{\includegraphics[width=4.8cm]{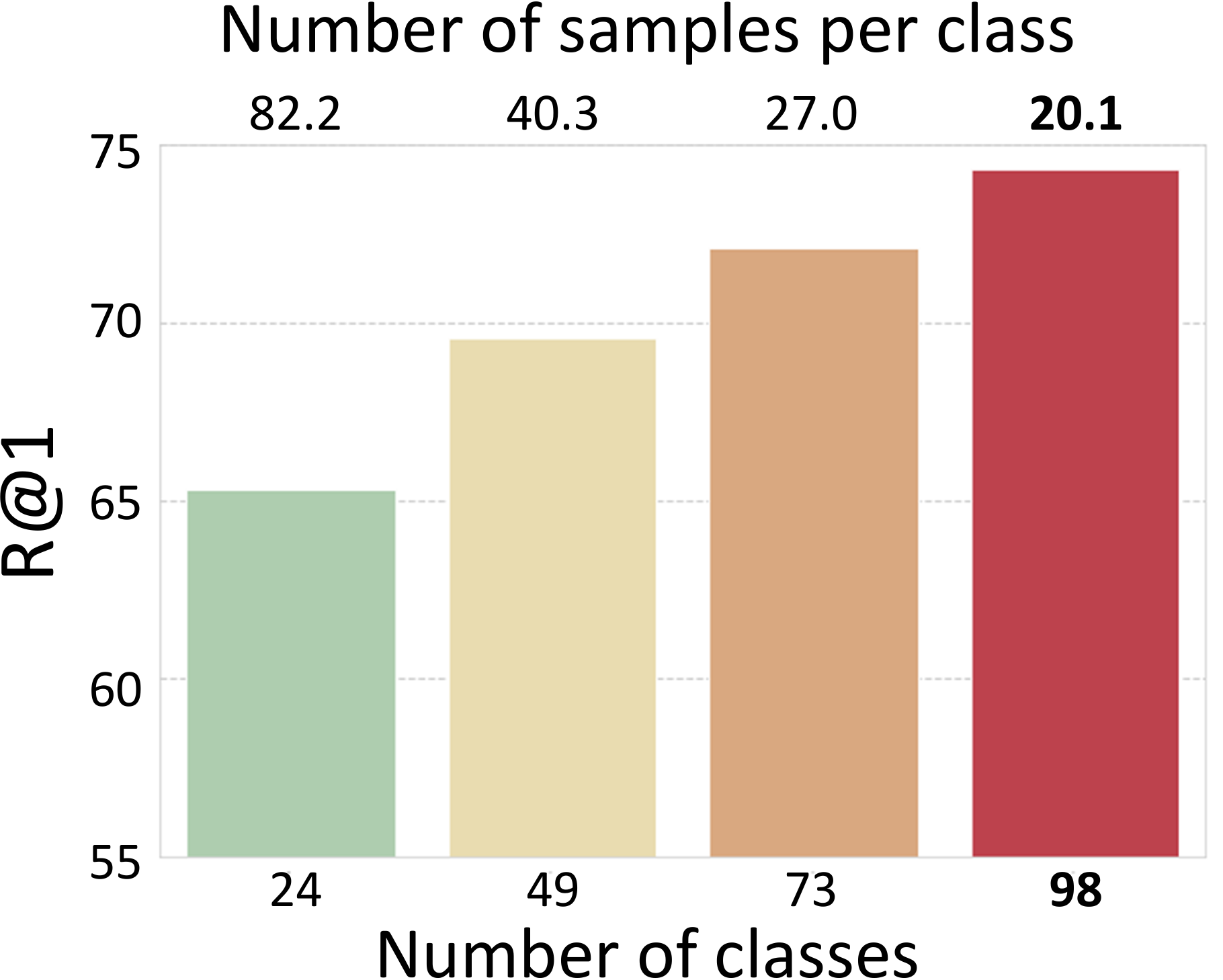}}&
\bmvaHangBox{\includegraphics[width=7.5cm]{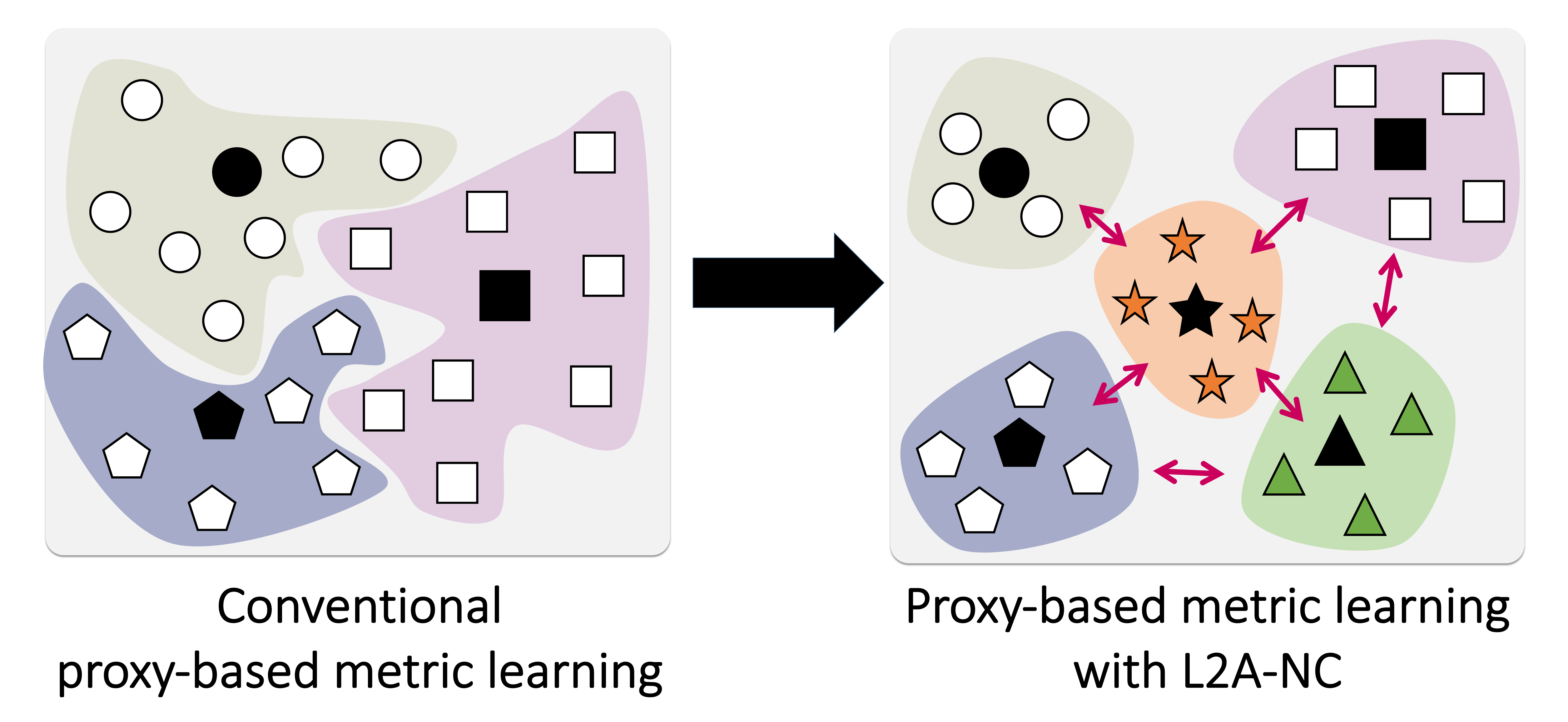}}
\end{tabular}
\caption{Our motivation and conceptual diagram. \textbf{Left}: Accuracy in Recall@1 versus the number of training classes with Proxy-Anchor loss on the Cars-196 dataset. The total number of samples used for training is fixed (Please see the first footnote.). \textbf{Right}: Comparison between proxy-based metric learning with and without L2A-NC. Black, empty, and colored nodes denote proxies, real embedding vectors, and synthetic embedding vectors of augmented classes, respectively. Also, different shapes indicate different classes.}
\label{fig:fig1}
\end{figure}

In this paper, we argue that the number of training classes is relatively more important than that of training samples in deep metric learning. We demonstrate the importance of the number of training classes by validating generalization ability while varying the number of training classes but fixing the total number of samples used in training. Specifically, for the test split of Cars-196 (\ie 8,131 images of the latter 98 classes), we measure the performances of models whose size of training classes ranges from 25$\%$ (24 classes) of total training classes to 100$\%$ (98 classes); reversely, the number of samples per training class decreases from 82.2 (for 25$\%$) to 20.1 (for 100$\%$) so that all models are trained with the same training sample size~\footnote{The fixed size of training data is 1,975 and \# of samples per training class is uniform.}. As shown in Figure~\ref{fig:fig1} (left), a larger number of classes lead to better performance although the number of samples per class becomes smaller. It is natural since more diverse classes would offer richer semantic relations between training classes. According to this observation, we remark the existing sample generation methods for deep metric learning are limited to leverage impoverished relations between given training classes.

Meanwhile, a recently proposed method~\cite{kim2020proxy} aims to synthesize new classes through linear interpolation between data representations of real classes so that inter-class relations with synthetic classes can yield better supervisory signals beyond relations between real classes. However, since this approach heavily relies on the data representations of real classes, the generated classes cannot cover a broad range of data characteristics, resulting in limitations in improving generalization.

In this paper, we introduce a novel data augmentation method that resolves the aforementioned issues. The key idea is to synthesize novel classes and their embedding vectors through a conditional generative model, which is an auxiliary module trained together with the main embedding network. We thus call it \emph{Learning to Augment Novel Classes}, dubbed \emph{L2A-NC}. Specifically, the conditional generative model, given a class label and a noise (\ie, a latent variable), produces a random embedding vector of the class. The model is trained by a metric learning loss like the main embedding network so that the novel classes and their embedding vectors become discriminative. At the same time, it is regularized to produce realistic embedding vectors by minimizing divergence between distributions of synthetic and real embedding vectors. As a result, novel classes have distributions that fit in between those of real classes in the learned embedding space (Figure~\ref{fig:fig1} (right)). The proposed method thus can synthesize realistic and discriminative novel classes thanks to the powerful expressiveness of deep neural networks trained with carefully designed loss functions. Consequently, these novel classes unavailable in original data help the main embedding network learn a better generalized embedding space. In summary, the contribution of this paper is three-fold:
\begin{itemize}
  \item We introduce a novel data augmentation framework for deep metric learning, called L2A-NC, which \emph{synthesizes novel classes and corresponding embedding vectors} to be augmented as additional training data through a conditional generative model.
  \item We design architecture and its training strategy that enable the generator to define novel classes and produce realistic and discriminative embedding vectors.
  \item It is demonstrated on public benchmarks that L2A-NC clearly enables non-trivial performance improvement of proxy-based losses.
\end{itemize}


\section{Related Work}
\label{sec:relatework}

\noindent\textbf{Proxy-based losses for deep metric learning.}
The loss functions for metric learning can be categorized into two types as \emph{pair-based} and \emph{proxy-based} losses. The pair-based losses are based on the pairwise relations between data in the embedding space. However, they have high training complexity since the number of tuples increases exponentially with the number of training data, forcing a careful tuple sampling technique. 
Proxy-based losses are proposed to alleviate the complexity issue by replacing samples with a small number of proxies, which are learnable parameters representing each class. Proxy-NCA~\cite{movshovitz2017no} is the first proxy-based method that pushes a sample to its positive proxy but repels against its negative proxies. Similarly, SoftTriple~\cite{Qian_2019_ICCV} assigns multiple proxies within one class to reflect intra-class variance. Proxy-Anchor~\cite{kim2020proxy} leverages data-to-data relations via forming a proxy as an anchor.

\noindent\textbf{Sample generation for deep metric learning.}
Sample generation methods are motivated to provide potentially informative samples which do not exist in the original data. \cite{Duan_2018_CVPR,Zheng_2019_CVPR,lin2018deep} exploit generative models to synthesize synthetic embedding vectors. To reduce training complexity, \cite{ko2020embedding,gu2020symmetrical} are proposed to generate synthetic embedding vectors by simple algebraic operation in the embedding space. However, these techniques are only coupled with pair-based losses and limited to synthesizing embedding vectors of existing classes.

\noindent\textbf{Virtual class synthesis.} Recently, several approaches have been proposed to utilize virtual classes in various areas. Virtual softmax~\cite{chen2018virtual} injects additional weight as a virtual negative class for softmax function. Proxy Synthesis~\cite{gu2020proxy} exploits virtual classes synthesized by linear interpolation between data representations of real classes. Different from them, our method synthesizes novel classes by a generative model, which utilizes the expressiveness of deep neural networks. In addition, most recently, VirFace~\cite{li2021virface} has been proposed to exploit unlabeled data as samples of virtual classes. Similar to our method, this approach introduces a VAE network to generate instances of virtual classes, but it also requires additional unlabeled data to train a generative model.

\section{Our Approach}
\label{sec:method}

\begin{figure*}[t]
    \centering
    \includegraphics[width=1.0\textwidth]{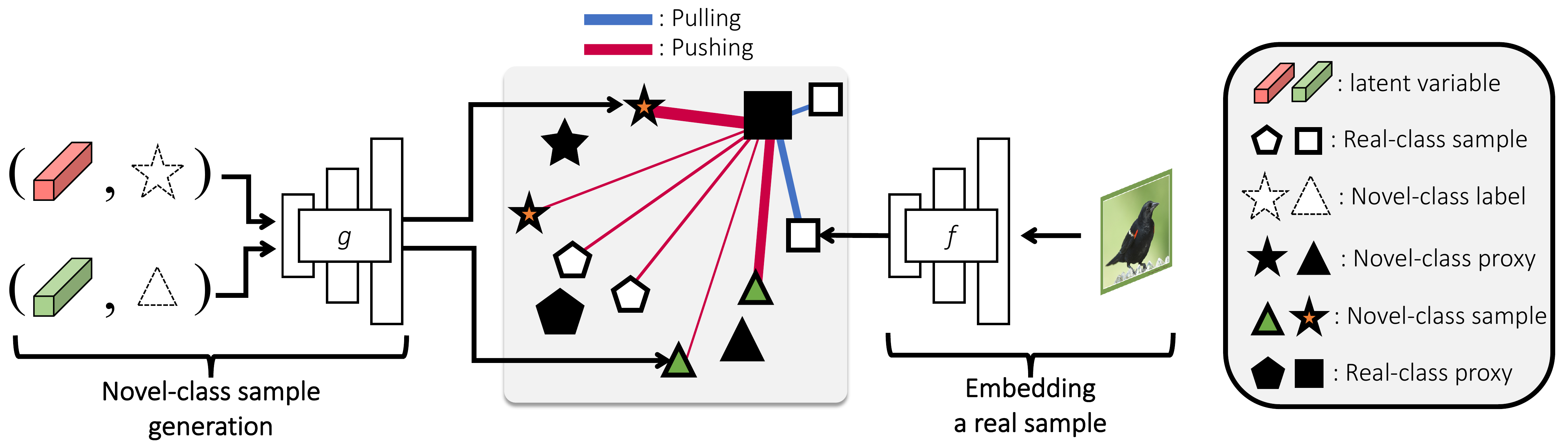}
    \caption{
    The overall framework of L2A-NC. 
    Given a novel class label and a latent variable, the conditional generative model $g$ produces an embedding vector of the class while the main model $f$ computes that of training images.
    In our framework, a proxy-based loss takes both of real and synthetic embedding vectors (Please refer Sec.~\ref{sec:method} for details). 
    }
    \label{fig:fig2}
\end{figure*}

As a way to improve the generalization of deep metric learning, we propose a new data augmentation method called L2A-NC, which synthesizes novel classes and their embedding vectors.
Our method learns and utilizes a conditional generative network that models novel classes and produces their embedding vectors, which are incorporated with any proxy-based losses to help them construct a more discriminative embedding space. The overall framework of L2A-NC is illustrated in Figure.~\ref{fig:fig2}.
In the rest of this section, we review the proxy-based metric learning losses, present details of the conditional generator, describe the training procedure incorporating L2A-NC, and analyze the effectiveness of novel classes with comparison to an existing class augmentation method.

\subsection{Background: Proxy-based Losses}

Suppose that we aim to learn an embedding network $f$ parameterized by $\theta_{f}$ and learnable proxies $P=[ p_{1}, ..., p_{C} ]$.
Let $X = [ x _{1}, ..., x_{N} ]$ be embedding vectors (\ie, the outputs of $f$) and $ Y = [ y_{1}, ..., y_{N}]$ be their corresponding labels, where $ y_{i} \in{ \{ 1, ..., C \} } $. 
A proxy-based loss optimized with respect to $\theta_{f}$ and $P$ is denoted by $J_{met}( X, Y, P)$.

In practice, however, proxy-based losses can be further enhanced in the sense that they are inherently limited only to training classes.
From this perspective, this paper proposes a new data augmentation method to synthesize novel classes and their embedding vectors. 

\subsection{Conditional Generator}
\label{sec:c_gen}
To synthesize novel classes, we train a conditional generator $ g $ which produces embedding vectors $\tilde{X} = [\tilde{x}_{1}, ..., \tilde{x}_{M}] $ of novel classes given corresponding labels $\tilde{Y} = [\tilde{y}_{1}, ..., \tilde{y}_{M}] $, where $ \tilde{y}_{j} \in{ \{ C+1, ..., C+\tilde{C} \} } $, and latent variables $Z$ for the stochastic generation:
\begin{equation}
\tilde{X} = g(\tilde{Y}, Z).
\end{equation}
Besides, corresponding proxies of novel classes $ \tilde{P}=[ p_{C+1}, ..., p_{C+\tilde{C} } ] $ are learnable, thus, first randomly initialized and updated via a  specified proxy-based loss just as real-class proxies $P$. Along with this conditional generation process, it is necessary to regularize our generator to produce realistic embedding vectors which are not far from the distribution of real embedding vectors; this is in line with previous data augmentation methods for deep metric learning~\cite{Schroff2015,Duan_2018_CVPR,Zheng_2019_CVPR}.
Furthermore, novel classes have to be discriminative so that they become diverse and independent from each other  like real classes of the original dataset.

\vspace{1mm}
\noindent \textbf{Loss functions.}
To guarantee that our generator produces realistic and discriminative embedding vectors, we introduce two loss functions for its training: A divergence loss $J_{div}$ and a proxy-based loss $J_{met}$. First, the generator is encouraged to fit its generated distributions in between those of real embedding vectors by minimizing the divergence loss.

As the Wasserstein distance has demonstrated its effectiveness in generative models~\cite{genevay2018learning,martin2017wasserstein} and other applications~\cite{zhou2020learning,liu2020semantic}, it is a good candidate for the divergence loss defined as:
\begin{equation}
\mathcal{W}(p,q)= \inf_{\gamma \in \Pi_{(p,q)}} \mathbb{E}_{x_{p},\ x_{q} \sim \gamma } [c(x_{p},x_{q}) ],
\label{eq:w_dist}
\end{equation}
where $ \Pi{}(p,q) $ is the set of all joint distributions $ \gamma{}(x_{p},x_{q}) $ and $c(\cdot,\cdot)$ denotes a cost function. This distance is usually interpreted as the minimum cost to turn the distribution $ q $ into the distribution $ p $. However, since the optimization problem in Eq.~\eqref{eq:w_dist} is generally intractable, 
we resort to the entropy-regularized Sinkhorn distance~\cite{cuturi2013sinkhorn}. In addition, to evaluate the Wasserstein distance on given mini-batches of $X_{p}$, $X_{q}$, we choose \textit{Sinkhorn AutoDiff}~\cite{genevay2018learning} proposed as an approximate of the distance:
\begin{equation}
\mathcal{W}_{c}( X_{p}, X_{q})=\inf_{M \in \mathcal{M} }[M \odot  C  ],
\end{equation}
where the cost function $c$ becomes the cost matrix $C$, where $C_{i,j}=c(x^{p}_{i}, x^{q}_{j})$, and the coupling distribution $ \gamma $ becomes the soft matching matrix $M$ whose all rows and columns sum to one. Although it is able to perform efficient optimization on GPUs, its gradients become no longer an unbiased gradient estimator when using mini-batches. Therefore, we finally adopt \textit{Mini-batch Energy Distance}~\cite{salimans2018improving}, which results in unbiased mini-batch gradients, as the divergence loss which is given by
\begin{equation}
\begin{aligned}
J_{div}(X, \tilde{X}) = & 2\mathbb{E}[\mathcal{W}_{c}(X_{1}, \tilde{X}_{1}) ] - \mathbb{E}[\mathcal{W}_{c}(X_{1}, X_{2})] -\mathbb{E}[\mathcal{W}_{c}(\tilde{X}_{1}, \tilde{X}_{2})],
\end{aligned}
\label{eq:j_div}
\end{equation}
where $X$ divided into $ X_{1} $ and $ X_{2} $ is a mini-batch from real data and $\tilde{X}$ divided into $\tilde{X}_{1} $ and $\tilde{X}_{2} $ is a mini-batch from generated data. For a cost function $c$, we adopt the cosine distance.

Second, we train the generator to produce discriminative embedding vectors. 
To this end, the generator aims to minimize a proxy-based loss that takes not only novel-class data but real-class data so that novel classes become diverse and offer richer class relations to an embedding model. Formally, the objective is given by 

\begin{equation}
J_{met}( X\cup\tilde{X}, Y\cup\tilde{Y}, P\cup\tilde{P}).
\label{eq:j_met}
\end{equation}

\subsection{Proxy-based Metric Learning with L2A-NC}
This section illustrates the overall pipeline of our method. 
We first pretrain the embedding function $f$ alone via a specific proxy-based loss $J_{met}$:

\begin{equation}
\min_{\theta_{f},P}J_{met}( X, Y, P).
\label{eq:eqn9}
\end{equation}
Then, we pretrain the conditional generator to optimize $J_{div}(X, \tilde{X})$ and $J_{met}(\tilde{X}, \tilde{Y}, \tilde{P})$ in advance since it is difficult for the generator to synthesize realistic and discriminative embedding vectors from scratch.
Finally, in the joint training phase, the two networks $f$ and $g$ are learned by optimizing the following common objective: 
\begin{equation}
\min_{\theta_{f},\theta_{g},P,\tilde{P}}J_{met}( X\cup\tilde{X}, Y\cup\tilde{Y}, P\cup\tilde{P}) +  \lambda_{div} J_{div}( X, \tilde{X}),
\label{eq:eqn11}
\end{equation}
where $\lambda_{div} $ is a hyperparameter to balance the two losses. Note that $J_{div}$ is optimized with respect to $\tilde{X}$ only, and encourages the generator to produce realistic embedding vectors in the joint training phase also. The overall training pipeline of L2A-NC is summarized in Section 1 of the supplementary material.

\begin{figure}
\includegraphics[width=0.98\textwidth]{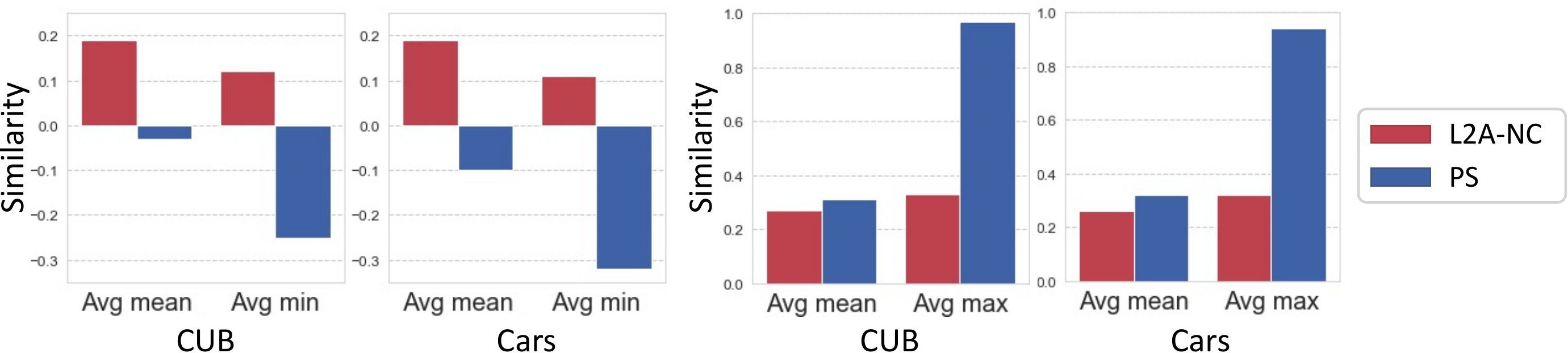}
    \caption{Cosine-similarity comparison. \textbf{Left}: One between embedding vectors and proxies of the same novel class. \textbf{Right}: One between proxies of real classes and those of novel classes.}
\label{fig:ps_ours}
\end{figure}

\subsection{Analysis of L2A-NC}
\label{sec:method_analysis}
In this section, we briefly review Proxy Synthesis (PS)~\cite{gu2020proxy}, an existing class augmentation method. Next, we analyze and compare the effectiveness of novel classes from PS and ours.

\vspace{1mm}
\noindent \textbf{Review of Proxy Synthesis (PS).}
As an existing method, PS synthesizes a synthetic proxy and a synthetic embedding vector by linear interpolation between proxies of different real classes, and embedding vectors of different real classes, respectively as

\begin{equation}
(\tilde{p}, \tilde{x}) = ( I_{\lambda_{ps}}(p_{i}, p_{j}), I_{\lambda_{ps}}(x_{i}, x_{j}) )
\label{eq:ps1}
\end{equation}
where $ y_{i} \neq y_{j} $, $ \tilde{x} \in \tilde{X} $, $ \tilde{p} \in \tilde{P} $,  and $ I_{\lambda_{ps}}(a, b) = \lambda_{ps}a+(1-\lambda_{ps}b) $ is a linear interpolation function with $ \lambda_{ps} \sim Beta(\alpha, \alpha) $ for $ \alpha \in (0, \infty) $, and $ \lambda_{ps} \in [0, 1] $. 

\vspace{1mm}
\noindent \textbf{Comparison to PS on the validity of novel classes.}
As previously discussed, learning with diverse classes improves performance as they allow to provide richer semantic relations. 
In this context, we verify that the proposed method can generate semantic and diverse classes like real classes, and compare it with PS\footnote{We adapt the official code from https://github.com/navervision/proxy-synthesis}. To this end, suppose $s(v_{i}, v_{j})$ denotes the cosine-similarity between two vectors, $v_{i}$ and $v_{j}$. Let $\tilde{x}_{i}$ and $\tilde{p}_{i} = p_{C+i}$ be an embedding vector and proxy of an arbitrary novel class label $\tilde{y}_{i}$. Next, we consider two cosine-similarities: one between embedding vectors and a proxy of the same novel class and another between proxies of real classes and those of novel classes (\ie $s( \tilde{x}_{i}, \tilde{p}_{i} )$ and $s(p_{j}, \tilde{p}_{i} )$, $\forall i \in \{1, ..., \tilde{C} \}, \forall j \in \{1, ..., C \}$). As shown in Figure~\ref{fig:ps_ours} (left), L2A-NC clearly shows high values of $s( \tilde{x}_{i}, \tilde{p}_{i} )$ while PS shows negative values on both mean and minimum on average. This suggests that L2A-NC generates novel classes that better preserve semantic properties while classes generated by PS fail to preserve their own semantics. Figure~\ref{fig:ps_ours} (right) shows that PS shows higher values of $s(p_{j}, \tilde{p}_{i} )$ than L2A-NC on both mean and maximum on average. This suggests that PS synthesizes classes that are highly redundant to real classes and lead to limited signals while L2A-NC generates diverse classes which provide richer semantic relations.

\section{Experiments}
\label{sec:experiments}
\vspace{-1mm}
In this section, to demonstrate the superiority of our framework, we compare L2A-NC with state-of-the-art methods and provide an in-depth analysis. Especially, we remark that L2A-NC also can be seamlessly incorporated with pair-based losses. Therefore, we further evaluate L2A-NC on pair-based losses as well as proxy-based losses.

\begin{table*}[!t]
    \resizebox{\textwidth}{!}{ %
    \begin{tabular}{ll| 
    ccc | 
    ccc | 
    ccc | 
    ccc
    }
    \hline
    
    \multicolumn{2}{l|}{} &
    \multicolumn{3}{c|}{CUB} & \multicolumn{3}{c}{Cars} & \multicolumn{3}{|c}{SOP} & \multicolumn{3}{|c}{In-Shop}
    \\ \cline{3-14}
    
    \multicolumn{1}{l}{Method} & \multicolumn{1}{c|}{Batch} & R@1 & R@2 & R@4 & R@1 & R@2 & R@4 & R@1 & R@10 & R@100 & R@1 & R@10 & R@20 
    \\ 
    \hline
    
    \multicolumn{1}{l}{Norm-softmax} & \multicolumn{1}{c|}{128} & 64.9 & 76.0 & 84.3 & 83.3 & 89.7 & 94.1 & 78.6 & 90.5 & 96.0 & 90.4 & 97.7 & 98.5
    \\
    
    \multicolumn{1}{l}{+ PS~\cite{gu2020proxy}} & \multicolumn{1}{c|}{128} & 66.0 & 76.6 & 85.0 & 84.7 & 90.7 & 94.6 & \textbf{79.6} & 90.9 & \textbf{96.2} & 91.5 & 98.1 & 98.7 \\
    
    \multicolumn{1}{l}{+ L2A-NC} & \multicolumn{1}{c|}{128} & \textbf{66.8}  & \textbf{77.0} & \textbf{85.6} & \textbf{86.0} & \textbf{91.8} & \textbf{95.2} & 79.4 & \textbf{91.0} & \textbf{96.2} & \textbf{91.9} & \textbf{98.2} & \textbf{98.8} \\
    \hline
    
    \multicolumn{1}{l}{Cosface~\cite{wang2018cosface}} & \multicolumn{1}{c|}{128} & 65.7 & 76.2 & 84.7 & 83.6 & 89.9 & 94.2 & 78.6 & 90.4 & 95.8 & 90.7 & 97.6 & 98.3 \\
    
    \multicolumn{1}{l}{+ PS~\cite{gu2020proxy}} & \multicolumn{1}{c|}{128} & 66.6 & 76.8 & 84.6 & 84.6 & 90.8 & 94.3 & \textbf{79.3} & 90.7 & 95.9 & 91.4 & 97.8 & 98.5 \\
    
    \multicolumn{1}{l}{+ L2A-NC} & \multicolumn{1}{c|}{128} & \textbf{67.6}  & \textbf{77.5} & \textbf{85.3} & \textbf{85.2} & \textbf{90.8} & \textbf{94.7} & \textbf{79.3} & \textbf{91.0} & \textbf{96.2} & \textbf{91.9} & \textbf{98.2} & \textbf{98.7} \\
    \hline
    
    \multicolumn{1}{l}{Proxy-NCA~\cite{movshovitz2017no}} & \multicolumn{1}{c|}{128} & 65.1 & 76.1 & 85.0 & 83.7 & 90.4 & 94.1 & 78.1 & 90.0 & 95.9 & 90.0 & 97.7 & 98.4 \\
    
    \multicolumn{1}{l}{+ PS~\cite{gu2020proxy}} & \multicolumn{1}{c|}{128} & 66.4 & 76.8 & 85.1 & 84.5 & 90.8 & 94.4 & 79.1 & 90.6 & 95.9 & 91.4 & 98.0 & 98.7 \\
    
    \multicolumn{1}{l}{+ L2A-NC} & \multicolumn{1}{c|}{128} & \textbf{67.7}  & \textbf{77.9} & \textbf{86.1} & \textbf{85.9} & \textbf{91.9} & \textbf{95.3} & \textbf{79.3} & \textbf{91.0} & \textbf{96.3} & \textbf{91.7} & \textbf{98.3} & \textbf{98.9} \\
    \hline
    
    \multicolumn{1}{l}{SoftTriple~\cite{Qian_2019_ICCV}$\dagger$} & \multicolumn{1}{c|}{128} & 66.3 & 76.8 & 84.7 & 84.9 & 90.5 & 94.3 & 79.0 & 90.7 & 96.1 & 91.1 & 97.8 & 98.4 \\
    
    \multicolumn{1}{l}{+ PS~\cite{gu2020proxy}} & \multicolumn{1}{c|}{128} & 66.6 & 76.8 & 85.1 & 85.3 & 91.0 & 94.8 & \textbf{79.5} & 90.6 & 96.0 & \textbf{91.8} & 98.1 & \textbf{98.7} \\
    
    \multicolumn{1}{l}{+ L2A-NC} & \multicolumn{1}{c|}{128} & \textbf{68.0} & \textbf{78.0} & \textbf{85.4} & \textbf{86.0} & \textbf{91.4} & \textbf{95.0} & 79.4 & \textbf{91.1} & \textbf{96.3} & 91.6 & \textbf{98.2} & \textbf{98.7} \\  
    \hline
    
    \multicolumn{1}{l}{Proxy-Anchor~\cite{kim2020proxy}$\ddag$} & \multicolumn{1}{c|}{180} & 69.1 & 78.9 & 86.1 & 86.4 & 91.9 & 95.0 & 79.2 & 90.7 & 96.2 & 91.9 & 98.1 & 98.7 \\
    
    \multicolumn{1}{l}{+ PS~\cite{gu2020proxy}} & \multicolumn{1}{c|}{180} & 69.2 & \textbf{79.5} & \textbf{87.2} & 86.9 & 92.4 & 95.2 & 79.8 & 90.9 & \textbf{96.4} &  91.9 & 98.2 & \textbf{98.8} \\
    
    \multicolumn{1}{l}{+ L2A-NC} & \multicolumn{1}{c|}{180} & \textbf{69.7} & 79.1 & 86.4 & \textbf{87.9} & \textbf{92.8} & \textbf{95.4} & \textbf{79.9} & \textbf{91.2} & 96.1 & \textbf{92.3} & \textbf{98.3} & 98.7 \\
    \hline
    
    \multicolumn{1}{l}{Average boost} &
    \multicolumn{1}{c|}{PS~\cite{gu2020proxy}} & \textcolor{forestgreen}{(+0.7)} & \textcolor{forestgreen}{(+0.5)} & \textcolor{forestgreen}{(+0.4)} &
    \textcolor{forestgreen}{(+0.8)} & \textcolor{forestgreen}{(+0.7)} & \textcolor{forestgreen}{(+0.3)} &
    \textbf{\textcolor{forestgreen}{(+0.8)}} & \textcolor{forestgreen}{(+0.3)} & \textcolor{forestgreen}{(+0.1)} & 
    \textcolor{forestgreen}{(+0.8)} & \textcolor{forestgreen}{(+0.3)} &
    \textcolor{forestgreen}{(+0.2)} \\
    
    \multicolumn{1}{l}{} &
    \multicolumn{1}{c|}{L2A-NC} & \textbf{\textcolor{forestgreen}{(+1.7)}} & \textbf{\textcolor{forestgreen}{(+1.1)}} & \textbf{\textcolor{forestgreen}{(+0.8)}} &
    \textbf{\textcolor{forestgreen}{(+1.8)}} & \textbf{\textcolor{forestgreen}{(+1.3)}} & \textbf{\textcolor{forestgreen}{(+0.8)}} &
    \textbf{\textcolor{forestgreen}{(+0.8)}} & \textbf{\textcolor{forestgreen}{(+0.6)}} & \textbf{\textcolor{forestgreen}{(+0.2)}} &
    \textbf{\textcolor{forestgreen}{(+1.0)}}  & \textbf{\textcolor{forestgreen}{(+0.5)}} & \textbf{\textcolor{forestgreen}{(+0.3)}} \\
    
    \hline
    \end{tabular} %
    }
    \vspace{-2mm}
    \caption{
        Comparison with the state-of-the-art methods. Image retrieval performance is measured as Recall@K ($\%$) on the public benchmark datasets. $\dagger$: For a fair comparison, we reproduced SoftTriple with the batch size of 128 using the author's official code and replace the original SoftTriple whose batch size is 32. $\ddag$: It is reported by the authors.  
    }
    \vspace*{-2mm}
    \label{tab:quant_proxy}
\end{table*}
\begin{table*}[!t]
    \centering
    \resizebox{\textwidth}{!}{
    \begin{tabular}{l | c c | c c c | c c | c c c}
    \hline
    
    \multicolumn{1}{l|}{} &
    \multicolumn{5}{c|}{CUB} & \multicolumn{5}{c}{Cars} \\ 
    \cline{1-11}
    
    \multicolumn{1}{l|}{Method} & NMI & F1 & R@1 & R@2 & R@4 & NMI & F1 & R@1 & R@2 & R@4 \\
    \hline
    
    \multicolumn{1}{l|}{Triplet$\dagger$} & 58.1 & 24.2 & 48.3 & 61.9 & 73.0 & 57.4 & 22.6 & 60.3 & 73.4 & 83.5 \\
    \multicolumn{1}{l|}{+ EE~\cite{ko2020embedding}}& \textbf{60.5} & \textbf{27.0} & 51.7 & 63.5 & 74.5 & \textbf{63.1} & \textbf{32.0} & 71.6 & 80.7 & 87.5 \\
    \multicolumn{1}{l|}{+ PS~\cite{gu2020proxy}}& 58.1 & 24.8 & 50.9 & 62.0 & 72.8 & 57.9 & 24.0 & 62.8 & 73.8 & 82.3 \\
    
    \multicolumn{1}{l|}{+ L2A-NC} & 59.4 & 26.0 & \textbf{53.6} & \textbf{65.3} & \textbf{75.6} & 61.6 & 29.2 & \textbf{73.0} & \textbf{81.9} & \textbf{88.2} \\
    \hline
    
    \multicolumn{1}{l|}{MS~\cite{wang2019multi}} & 62.8 & 31.2 & 56.2 & 68.3 & 79.1 & 62.4 & 30.2 & 75.0 & 83.1 & 89.5 \\
    \multicolumn{1}{l|}{+ EE~\cite{ko2020embedding}}& 63.3 & 32.5 & 57.4 & 68.7 & 79.5 & 63.5 & 33.5 & 76.1 & 84.2 & 89.8 \\ 
    
    \multicolumn{1}{l|}{+ PS~\cite{gu2020proxy}}& 61.1 & 29.5 & 55.9 & 68.1 & 78.1 & 58.0 & 25.0 & 71.8 & 80.9 & 87.6 \\
    
    \multicolumn{1}{l|}{+ L2A-NC} & \textbf{66.1} & \textbf{35.8} & \textbf{60.6} & \textbf{72.5} & \textbf{82.2} & \textbf{68.1} & \textbf{37.9} & \textbf{81.2} & \textbf{88.4} & \textbf{93.0} \\
    \hline
    
    \multicolumn{1}{l|}{EE~\cite{ko2020embedding}}& \textcolor{forestgreen}{(+1.4)} & \textcolor{forestgreen}{(+2.0)} & \textcolor{forestgreen}{(+2.3)} & \textcolor{forestgreen}{(+1.0)} & \textcolor{forestgreen}{(+0.9)} & \textcolor{forestgreen}{(+3.4)} & \textcolor{forestgreen}{(+6.5)} & \textcolor{forestgreen}{(+6.1)} & \textcolor{forestgreen}{(+4.2)} & \textcolor{forestgreen}{(+2.1)} \\
    
    \multicolumn{1}{l|}{PS~\cite{gu2020proxy}}& \textcolor{red}{(-0.8)} & \textcolor{red}{(-0.5)} & \textcolor{forestgreen}{(+1.1)} & (0.0) & \textcolor{red}{(-0.6)} & \textcolor{red}{(-1.9)} & \textcolor{red}{(-1.9)} & \textcolor{red}{(-1.9)} & \textcolor{red}{(-0.3)} & \textcolor{red}{(-1.5)} \\
    
    \multicolumn{1}{l|}{L2A-NC} & \textbf{\textcolor{forestgreen}{(+2.3)}} & \textbf{\textcolor{forestgreen}{(+3.2)}} & \textbf{\textcolor{forestgreen}{(+4.9)}} & \textbf{\textcolor{forestgreen}{(+3.8)}} & \textbf{\textcolor{forestgreen}{(+2.9)}} & \textbf{\textcolor{forestgreen}{(+4.9)}} & \textbf{\textcolor{forestgreen}{(+7.1)}} & \textbf{\textcolor{forestgreen}{(+9.5)}} & \textbf{\textcolor{forestgreen}{(+6.9)}} & \textbf{\textcolor{forestgreen}{(+4.1)}} \\
    \hline
    \end{tabular}
    }
    \vspace{-2mm}
    \caption{
        Comparison with the existing pair-based losses. NMI and F1 ($\%$) are measured for clustering performance. Recall@K ($\%$) is measured for retrieval performance. $\dagger$ denotes the triplet loss with hard tuple mining.
    }
    \vspace*{-2mm}
    \label{tab:quant_pair}
\end{table*}

\begin{figure*}[!t]
    \centering
    \includegraphics[width=0.8\textwidth]{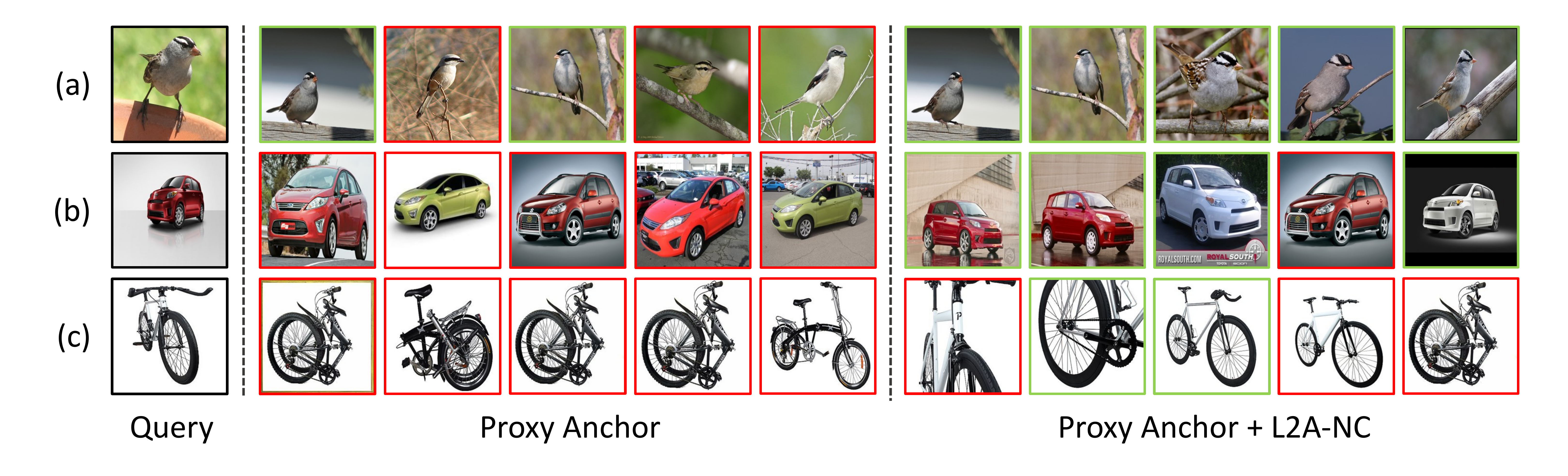}
    \vspace{-4mm}
    \caption{Qualitative results of the vanilla Proxy-Anchor~\cite{kim2020proxy} and that combined with L2A-NC on the (a) CUB, (b) Cars, and (c) SOP datasets. Images with green boundary are correct results while those with red boundary are failure cases. 
    }
    \label{fig:fig3}
\end{figure*}

\subsection{Setup}
\noindent \textbf{Datasets.} 
Combinations of L2A-NC and proxy-based losses are evaluated on benchmark datasets for deep metric learning: CUB-200-2011 (CUB)~\cite{CUB200-2011}, Cars-196 (Cars)~\cite{krause20133d}, Stanford Online Product (SOP)~\cite{songCVPR16}, and In-shop Clothes Retrieval (In-Shop)~\cite{DeepFashion}.
For splitting training and test sets, we directly follow the widely used setting in~\cite{songCVPR16}. For comparison with existing pair-based losses, CUB and Cars are adopted.

\vspace{1mm}
\noindent \textbf{Network architectures.} 
We adopt ImageNet pre-trained BatchNorm Inception~\cite{batchnorm} and GoogleNet~\cite{Googlenet} for experiments associated with proxy-based losses and pair-based losses, respectively. For all experiments, the dimensionality of embedding vectors is 512. The conditional generator consists of 4 fully connected layers; a conditional batch normalization layer~\cite{de2017modulating} is inserted between every pair of layers so that the class information is injected. Also, the dimension of input noise for the generator is fixed to 16 for all experiments.

\vspace{1mm}
\noindent \textbf{Details of training.} 
In the pretraining stage, the main embedding models are trained by directly following the setting (\eg batch size) presented in PS~\cite{gu2020proxy}, while the conditional generator is optimized by AdamW~\cite{adamw} with the learning rate of
$10^{-4}$ for all datasets.
In the joint training stage, the learning rate of the embedding network is set to $5^{-5}$ for all datasets. For the main embedding models incorporated with pair-based losses, we directly follow the setting in \cite{ko2020embedding} and apply our framework described above.

\subsection{Comparison with State of the Art}
\noindent \textbf{Results on proxy-based losses.} 
As summarized in Table~\ref{tab:quant_proxy}, we observe that L2A-NC incorporated with proxy-based losses achieves non-trivial performance boosts in all datasets. Especially, on the CUB and Cars datasets, our method outperforms the vanilla method and PS~\cite{kim2020proxy} by a non-trivial margin (by 1.7\%p and 1.0\%p, respectively, in average Recall@1.). However, on the SOP and In-Shop datasets, we find the tendency that performance boosts of L2A-NC decreases compared to those on the CUB and Cars datasets, though L2A-NC still shows competitive or better performance boosts than PS. We conjecture this is because the SOP and In-Shop datasets have already a lot of training classes (11,318 and 3,997, respectively) which are about 113 and 40 times compared to those of CUB or Cars. Nevertheless, compared to PS, L2A-NC achieves more performance boosts except for R@1 on the SOP.

\vspace{1mm}
\noindent \textbf{Results on pair-based losses.}
As shown in Table~\ref{tab:quant_pair}, L2A-NC outperforms not only the vanilla loss but also PS. We find that PS fails to boost pair-based losses in most cases even though it shows its effectiveness with proxy-based ones. Since pair-based losses are known to be more vulnerable to noisy labels or outliers than proxy-based losses, we conjecture it is because noisy synthetic classes and their embedding vectors of PS hamper model generalization from scratch, which is discussed in Sec.~\ref{sec:method_analysis}. Furthermore, even when compared to the current state-of-the-art sample generation method, Embedding Expansion (EE)~\cite{ko2020embedding}, dedicated to pair-based losses, our method achieves larger performance improvements. Note that no proxies are introduced in any procedure of ours and PS in Table~\ref{tab:quant_pair} for a fair comparison.

\begin{table}[!t]
\begin{minipage}[t]{0.48\linewidth}
\vspace{0.37mm}
\fontsize{6}{7.8}\selectfont
\centering
\begin{tabularx}{1\textwidth}{ 
   >{\centering\arraybackslash}X |
   >{\centering\arraybackslash}X 
   >{\centering\arraybackslash}X 
   >{\centering\arraybackslash}X 
   >{\centering\arraybackslash}X 
   >{\centering\arraybackslash}X}
\hline
\multicolumn{1}{l|}{Dataset} & +0\% & +25\% & +50\% & +100\% & +200\% \\ \hline
\multicolumn{1}{l|}{CUB} & 69.1 & 69.3 & 69.6 & 69.5 & \textbf{69.7} \\
\multicolumn{1}{l|}{SOP}& 79.2 & 79.3 & 79.5 & 79.5 & \textbf{79.9} \\
\multicolumn{1}{l|}{In-Shop} & 91.9 & 92.0 & 92.2 & 92.2 & \textbf{92.3} \\ \hline
\end{tabularx}
\end{minipage}
\hfill
\begin{minipage}[t]{0.5\linewidth}
\vspace{0.37mm}
\fontsize{6}{7.8}\selectfont
\centering
\begin{tabularx}{1\textwidth}{ 
   >{\centering\arraybackslash}X |
   >{\centering\arraybackslash}X
   >{\centering\arraybackslash}X 
   >{\centering\arraybackslash}X}
  \hline
\multicolumn{1}{l|}{Dataset}  & Vanilla & L2A-EC & L2A-NC (Ours)\\ \hline
\multicolumn{1}{l|}{CUB}  & 69.1  & 69.2 & \textbf{69.7} \\ 
\multicolumn{1}{l|}{Cars} & 86.4 & 86.8  & \textbf{87.9}\\ 
\multicolumn{1}{l|}{In-Shop} & 91.9 & 91.8  & \textbf{92.3}\\ \hline
\end{tabularx}
\end{minipage}
\vspace{0.5em}
\caption{Ablation studies on proposed L2A-NC. \textbf{Left}: Recall@1 versus the number of novel classes. \textbf{Right}: Comparison between L2A-EC and L2A-NC in Recall@1.}
\label{tab:tab2}
\end{table}

\begin{figure*}[t]
    \centering
    \includegraphics[width=0.8\columnwidth]{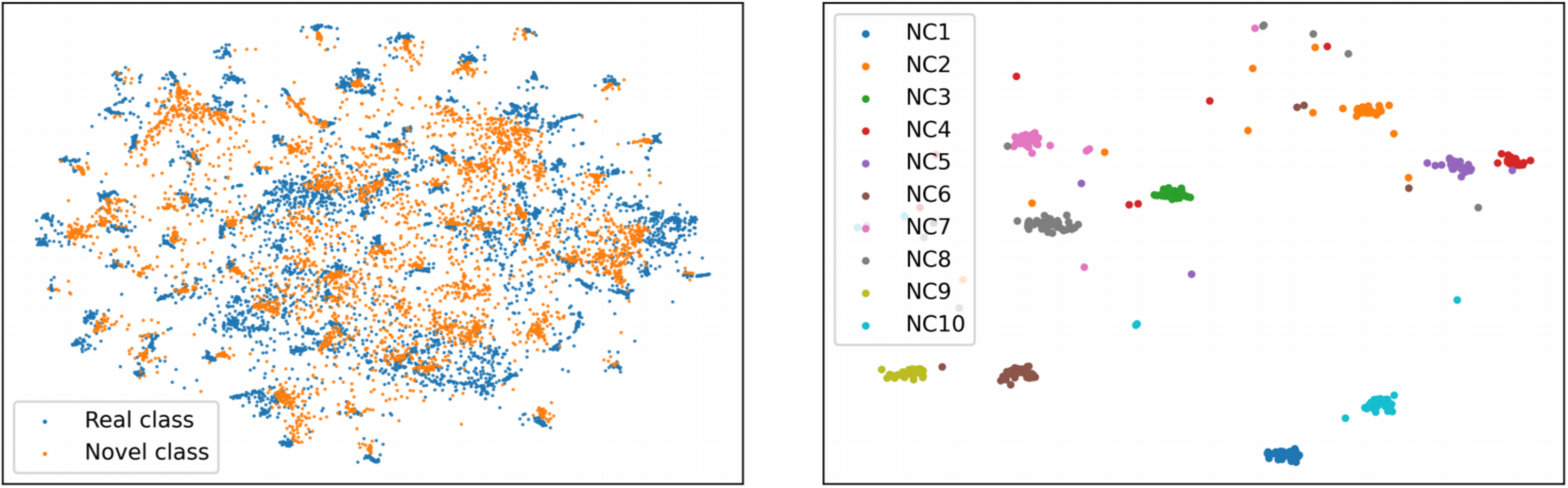}
   \vspace{-2mm}
   \caption{
   $t$-SNE visualizations of the learned embedding space. 
   \textbf{Left}: Embedding vectors of both real and novel classes. \textbf{Right}: Embedding vectors of novel classes only.
   }
    \label{fig:fig4}
\end{figure*}

\subsection{In-depth Analysis on L2A-NC}
\noindent \textbf{Image retrieval results.}
We further demonstrate the superiority of L2A-NC through qualitative results of image retrieval. 
Figure.~\ref{fig:fig3} presents image retrieval examples of Proxy-Anchor loss incorporating L2A-NC and those of the vanilla Proxy-Anchor. We observe that results of the vanilla Proxy-Anchor are mostly incorrect and biased towards backgrounds rather than target objects, whereas L2A-NC enables retrieval of correct images regardless of background. For example, the vanilla version failed to retrieve even a single correct image for the Cars and SOP. However, the version incorporating L2A-NC retrieved correct images although colors or viewpoints of target objects are substantially different from those of the query. We demonstrate that L2A-NC functions as desired to regularize the vanilla method not to be biased towards certain prevailing features in the dataset (\eg, background or viewpoint) by augmenting novel classes unavailable in the original data.

\vspace{1mm}
\noindent \textbf{Impact of the number of novel classes.}
Thanks to the architecture of our conditional generator, L2A-NC can define and synthesize an arbitrary number of novel classes. 
To investigate the impact of the number of novel classes, we evaluate the performance of L2A-NC incorporated with Proxy-Anchor loss  on the CUB, SOP, and In-Shop while varying the number of novel classes. 
As shown in Table~\ref{tab:tab2} (left), the performance increases by adding more novel classes to some degree, but there is an upper bound of the effectiveness; 
it is natural since too many novel classes may prevent the embedding model from understanding relations between classes, leading to unstable training. In experiments, we fixed the number of novel classes for all datasets: 200\% of the number of existing classes.

\vspace{1mm}
\noindent \textbf{Importance of generating novel classes.}
To verify that our performance boost is attributed to the augmentation of embedding vectors of novel classes, we compare L2A-NC with its variant that produces synthetic embedding vectors of \textit{existing classes}; we call it L2A-EC. L2A-NC and L2A-EC are incorporated with Proxy-Anchor loss and compared on the CUB and the Cars in terms of Recall@1. As shown in Table~\ref{tab:tab2} (right), compared to L2A-NC, L2A-EC provided smaller improvements and even slightly worsened the vanilla method. These results are consistent with our hypothesis that novel classes unavailable in the original dataset would provide richer signals than existing class data.

\begin{table}[!t]
\begin{minipage}[t]{0.46\linewidth}
\fontsize{6}{7.8}\selectfont
\centering
\begin{tabularx}{1\textwidth}{ 
   >{\centering\arraybackslash}X |
   >{\centering\arraybackslash}X}
\hline
\multicolumn{1}{c|}{Class pair}& Mean of minimum KL divergence \\ \hline
\multicolumn{1}{c|}{Train--Test}& 56.7\\ 
\multicolumn{1}{c|}{Novel--Test}& \textbf{27.9}\\ \hline
\end{tabularx}
\vspace{-1\baselineskip}
\caption{Quantitative analysis about how closely training and novel classes approximate unseen test classes on the CUB.}
\label{tab:tab3}
\end{minipage}
\hfill
\begin{minipage}[t]{0.52\linewidth}
\fontsize{6}{7.8}\selectfont
\centering
\begin{tabularx}{1\textwidth}{ 
   >{\centering\arraybackslash}X |
   >{\centering\arraybackslash}X |
   >{\centering\arraybackslash}X |
   >{\centering\arraybackslash}X}
  \hline
\multicolumn{1}{l|}{Method}  & sec / iter  &  \# of parameters & R@1 boost\\ \hline
\multicolumn{1}{l|}{Vanilla}  & 0.28  & 11.85M & -\\ 
\multicolumn{1}{l|}{PS~\cite{gu2020proxy}} & 0.37 & 11.85M & 1.1\\
\multicolumn{1}{l|}{Ours} & 0.43 & 12.24M & 1.9 \\ \hline
\end{tabularx}
\vspace{-1\baselineskip}
\caption{Performance boost over training complexity.}
\label{tab:tab4}
\end{minipage}
\end{table}

\vspace{1mm}
\noindent \textbf{\textit{t}-SNE visualization.}
To illustrate how embedding vectors of novel classes offer additional signals, we visualize the learned embedding space. As shown in Figure~\ref{fig:fig4}, the generator synthesizes embedding vectors that are realistic and discriminative. In detail, Figure~\ref{fig:fig4} (left) shows that embedding vectors of novel classes are located in between those of real classes with forming their own clusters where embedding vectors of real classes do not exist. In addition, Figure~\ref{fig:fig4} (right) shows how novel classes are discriminative to each other. With these results, we demonstrate how novel classes unavailable in the original dataset can offer additional signals, which lead to better generalization on unseen classes.

\vspace{1mm}
\noindent \textbf{Relation between novel classes and test classes.}
To demonstrate that the novel classes we generate affect the robustness on unseen classes, we investigate how well the novel and unseen test classes are aligned to each other in the learned embedding space.
Specifically, we quantify the degree of alignment through KL divergence: Each novel class is first matched with its nearest test class with the minimum KL divergence, then the minimum divergence values of all novel classes are averaged. We measure this score for training classes as well for comparison. The results in Table~\ref{tab:tab3} demonstrate that novel classes better approximate test classes than training classes even though the conditional generator of L2A-NC is not aware of test data; this suggests that the proposed method has the potential to improve the generalization of the learned embedding space. 

\vspace{1mm}
\noindent \textbf{Training complexity.}
To be a promising option for practical usage, it is quite desirable for L2A-NC to offer an appealing trade-off between performance boost and training complexity. As shown in Table~\ref{tab:tab4}\footnote{All the results were produced on a NVIDIA TITAN RTX GPU.}, compared to PS~\cite{gu2020proxy}, L2A-NC achieves about 1.7 times more performance boost averaged over every benchmark when incorporated with Proxy-NCA at the cost of only 16\% and 3\% increase in training time and parameters. This result is brought by our lightweight design: the generator consists of only a few layers and holding in memory novel-class proxies is also negligible as well.

\section{Conclusion}
\label{sec:conclusion}We have presented a novel data augmentation method for deep metric learning. 
Distinct from existing techniques, the proposed method synthesizes novel classes and their embedding vectors through a conditional generative model.
Thanks to the carefully designed loss functions and its architecture, the generator synthesizes novel classes that are realistic and discriminative so it can offer richer semantic relations to an embedding model. As a result, our method enabled both proxy-based and pair-based losses to improve the quality and search performance of the learned embedding space.


\section*{Acknowledgements}
This work was supported by 
the NRF grant,
the IITP grant,
and R\&D program for Advanced Integrated-intelligence for IDentification,
funded by Ministry of Science and ICT, Korea
(No.2019-0-01906 Artificial Intelligence Graduate School Program--POSTECH,
 NRF-2021R\\1A2C3012728--50\%,
 NRF-2018M3E3A1057306--50\%).

\bibliography{cvlab_kwak}
\end{document}